\documentclass[conference]{IEEEtran}
\IEEEoverridecommandlockouts

\usepackage[utf8]{inputenc}
\usepackage{url}
\usepackage{hyperref}

\usepackage{cite}
\usepackage{float}
\usepackage{amsmath,amssymb,amsfonts}
\usepackage{amsthm}

\usepackage{graphicx}
\usepackage{tikz}

\usepackage{tabu}
\usepackage{array}
\newcolumntype{P}[1]{>{\centering\arraybackslash}p{#1}}
\newcolumntype{M}[1]{>{\centering\arraybackslash}m{#1}}

\usepackage{algorithm}
\usepackage{algpseudocode}

\usepackage{caption}
\usepackage{subcaption}
\usepackage{textcomp}
\usepackage{xcolor}
\usepackage[english]{babel}
\usepackage{float}
\usepackage{nomencl}
\usepackage{multicol}
\usepackage{multirow}
\usepackage{makecell}

\def\BibTeX{{\rm B\kern-.05em{\sc i\kern-.025em b}\kern-.08em T\kern-.1667em\lower.7ex\hbox{E}\kern-.125emX}}



\begin{document}

\title{Enhancing Predictive Accuracy in Pharmaceutical Sales Through An Ensemble Kernel Gaussian Process Regression Approach}

\author{Shahin Mirshekari, Mohammadreza Moradi, Hossein Jafari, Mehdi Jafari, Mohammad Ensaf~\emph{Student Member, IEEE}}

% \author{\IEEEauthorblockN{1\textsuperscript{st} Mohammad Ensaf}
% \IEEEauthorblockA{\textit{dept. Electrical and Computer Engineering} \\
% \textit{University of Pittsburgh}\\
% Pittsburgh, United States\\
% mohammad.ensaf@pitt.edu} 
% \and

% \IEEEauthorblockN{2\textsuperscript{nd} Masoud Barati}
% \IEEEauthorblockA{\textit{dept. Electrical and Computer Engineering} \\
% \textit{University of Pittsburgh}\\
% Pittsburgh, United States \\
% masoud.barati@pitt.edu} 
% \and

        % <-this % stops a space
\thanks{Shahin Mirshekari is with the Department of Marketing Science and Business Analytics, Katz Graduate School of Business, University of Pittsburgh, PA, USA,  shm177@pitt.edu. Mohammadreza Moradi is with the Department of Biomedical Engineering, Carnegie Mellon University, PA, USA,  mmoradi2@cs.cmu.edu. Mohammad Ensaf and Hossein Jafari are with the Department of Electrical and Computer Engineering and Industrial Engineering, Swanson School of Engineering, University of Pittsburgh, PA, USA, {(mohammad.ensaf, hoj27@pitt.edu)}. Mehdi Jafari is with the Department of Business, Duquesne University, PA, USA,  jafarim@duq.edu. % <-this % stops a space
}

%\author{Masoud~Barati,~\IEEEmembership{Senior Member,~IEEE}
        % <-this % stops a space
%\thanks{M. Barati is with the Department
%of Electrical and Computer Engineering and Department of Industrial Engineering, University of Pittsburgh, Pittsburgh,
%PA, 15206 USA e-mail: masoud.barati@pitt.edu}% <-this % stops a space


\markboth{ }
\maketitle

\IEEEoverridecommandlockouts
\IEEEpubid{\makebox[\columnwidth]{} \hspace{\columnsep}\makebox[\columnwidth]{ }}
\maketitle
\begin{abstract}
This research employs Gaussian Process Regression (GPR) with an ensemble kernel, integrating Exponential Squared, Revised Matérn, and Rational Quadratic kernels to analyze pharmaceutical sales data. Bayesian optimization was used to identify optimal kernel weights: 0.76 for Exponential Squared, 0.21 for Revised Matérn, and 0.13 for Rational Quadratic. The ensemble kernel demonstrated superior performance in predictive accuracy, achieving an R² score near 1.0, and significantly lower values in MSE, MAE, and RMSE. These findings highlight the efficacy of ensemble kernels in GPR for predictive analytics in complex pharmaceutical sales datasets.
\end{abstract}


% \fi
\vspace{-5pt}
\IEEEpeerreviewmaketitle
\vspace{-5pt}
\section{Introduction}
\begin{figure}[!htb]
\centering
\includegraphics[width=0.5\textwidth]{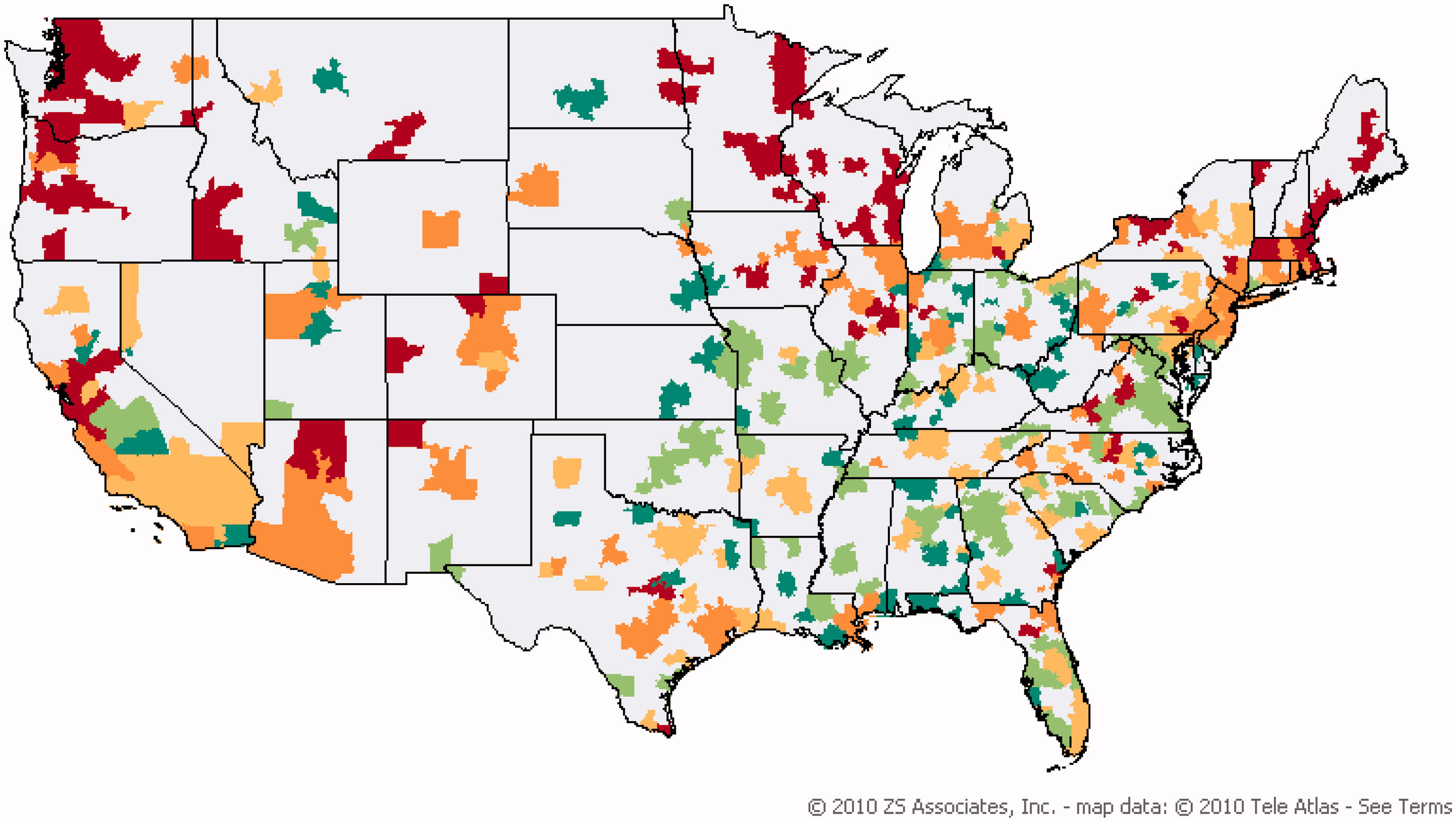}
\caption{Sales of Pharmaceutical Products Across Various States in the United States\cite{11}}
\label{fig_a}
\end{figure}

\IEEEPARstart{P}{h} armaceutical data and time series analysis are pivotal in the healthcare sector, offering profound insights into drug development, patient care, and market trends. In pharmaceutical data, information ranges from drug efficacy and safety profiles to patient health outcomes and market dynamics. Time series analysis, a method of analyzing data points collected or recorded at regular time intervals, plays a crucial role in understanding these aspects. It allows researchers and healthcare professionals to observe trends, seasonal patterns, and long-term changes in pharmaceutical data, enabling them to make data-driven decisions. This approach is particularly valuable in monitoring drug performance over time, understanding patient response to treatments, and predicting future market needs. Fig.\ref{fig_a} illustrates the variation in pharmaceutical sales across different states in the United States, highlighting the geographical differences in market dynamics. Time series analysis in pharmaceuticals is not just about handling large volumes of data; it's about extracting meaningful patterns and insights that can lead to more effective treatments and strategies in healthcare \cite{1,2,3,4,5,6,7}.

Analyzing time series in pharmaceutical sales data presents distinct challenges, underscored by the complexity and volatility of the market. A primary challenge lies in accommodating the unpredictable fluctuations in demand, influenced by factors like market trends, regulatory changes, and competitive dynamics. The uncertainty is further amplified by external variables such as healthcare policies, economic conditions, and public health crises, which can drastically impact sales patterns. Another hurdle is the seasonality of certain medications, where sales peak during specific times of the year, requiring sophisticated models to accurately forecast trends\cite{8,9,10,11,12}.

Additionally, data inconsistency and incompleteness, often due to varying reporting standards across regions and channels, can complicate the analysis. These challenges necessitate advanced analytical techniques and robust data-handling strategies to extract reliable insights and forecasts from pharmaceutical sales time series.

Time series analysis in the pharmaceutical industry, particularly in sales data, has been a topic of interest in both classic and modern literature. Classic methods often involve statistical models such as Autoregressive Integrated Moving Average (ARIMA), Seasonal Autoregressive Integrated Moving Average (SARIMA), and Autoregressive Moving Average (ARMA). These models are renowned for their proficiency in identifying linear sequences in time-series data. Yet, their predictive accuracy might be compromised due to presumptions of immediate reactions. Conversely, contemporary approaches have embraced machine learning and deep learning methodologies. For example, employing both basic and complex neural networks for anticipating demand has been a recent trend. These models focus on devising strategies for sales and marketing by assessing the periodic and trend-related influences on different categories of pharmaceutical items. Research has demonstrated that Demand Forecast Models (DFMs) grounded in basic neural networks are capable of precisely forecasting future needs for pharmaceutical products. Furthermore, innovative combined neural network models have been crafted to effectively track both straightforward and complex patterns in sales data \cite{3}. Other modern methods include Holt-Winter Exponential Smoothing, Linear Regression, Artificial Neural Network, and XGBoost. Both classic and modern methods of time series analysis play crucial roles in the pharmaceutical industry, particularly in sales data forecasting. They provide valuable insights that can help companies make informed decisions and strategies\cite{13,14,15,16,17,18}.

In this paper, we aim to harness the power of Bayesian models and Gaussian Process Regression (GPR) for time series analysis. Bayesian models are particularly advantageous as they provide a formal way to incorporate prior information, fit perfectly with sequential learning and decision-making, and lead to exact small sample results. They also allow us to take advantage of the association structure among target series, select important features, and train the data-driven model simultaneously. GPR, on the other hand, is a powerful tool for modeling correlated observations, including time series. It provides a prior over functions, capturing prior beliefs about the function behavior, such as smoothness or periodicity. This allows us to perform robust modeling even in highly uncertain situations. These methods are well-suited to the dynamic, noisy, and time-sensitive environments often encountered in time series analysis.

In this paper, our contributions are threefold. First, we utilize \textbf{Gaussian Process Regression (GPR)} as a powerful tool for time series analysis. Second, we employ \textbf{different kernels} and introduce an \textbf{ensemble kernel}, which combines the strengths of individual kernels to capture more complex patterns in the data. Third, we leverage \textbf{Bayesian optimization} to find the optimal weights of the kernels in the ensemble kernel in GPR, thereby fine-tuning our model to the specific characteristics of the data. Finally, we evaluate our approach using general metrics, demonstrating that our \textbf{ensemble method outperforms the individual kernels}.

\subsection{Neural Networks vs. Gaussian Processes in Time Series Analysis}

In the context of time series analysis, the distinction between neural networks and Bayesian methods like Gaussian processes becomes particularly pronounced. Neural networks, especially recurrent neural networks (RNNs) and their variants like Long Short-Term Memory (LSTM) networks, are highly effective for time series forecasting. They excel in capturing complex temporal dependencies and patterns in large datasets, making them ideal for applications like stock market prediction or weather forecasting. Conversely, Gaussian processes, embodying a Bayesian approach, excel in modeling time series data where uncertainty quantification is crucial. They provide a probabilistic forecast and are particularly valuable in scenarios with sparse or irregularly sampled data. Gaussian processes are also favored for their ability to incorporate prior knowledge about the time series through the kernel function, offering a more interpretable model compared to the often "black-box" nature of deep neural networks. Thus, while neural networks are suited for large-scale, complex temporal pattern modeling, Gaussian processes offer robust and interpretable solutions for time series forecasting, especially when dealing with uncertainty and requiring insights into the model's behavior.

\subsection{Detailed Exposition of Gaussian Processes}

Gaussian Processes (GPs), integral to machine learning, especially in regression and classification, are particularly valuable for analyzing time series. Defined as a collection of random variables, where any finite number have a joint Gaussian distribution, GPs are characterized by two primary components: a mean function, typically denoted \( \mu(x) \), and a covariance function, or kernel, represented as \( \kappa(x, x') \).

\begin{equation}
g(x) \sim GP(\mu(x), \kappa(x, x'))
\label{eq:gp_model}
\end{equation}

Here, \( g(x) \) symbolizes the Gaussian Process, with \( GP \) indicating a Gaussian Process. The mean function \( \mu(x) \) is often assumed to be zero, as GPs can encapsulate the mean adequately through the covariance function. The kernel \( \kappa(x, x') \) is crucial as it determines how similar input points are considered in the process.

Kernels like the Squared Exponential, Matérn, and Rational Quadratic are pivotal in GP models. They encode assumptions about the target function and are selected based on the data characteristics and desired function properties.

For a dataset \( \mathcal{D} = \{ (x_i, y_i) \}_{i=1}^n \) with \( y_i \) as observations and \( x_i \) as inputs, the joint distribution of observed targets and function values at a new point \( x^* \) is modeled as:

\begin{equation}
\begin{pmatrix}
\mathbf{y} \\
g(x^*)
\end{pmatrix}
\sim 
\mathcal{N} \left(
\mathbf{0},
\begin{pmatrix}
\kappa(X, X) + \epsilon^2I & \kappa(X, x^*) \\
\kappa(x^*, X) & \kappa(x^*, x^*)
\end{pmatrix}
\right)
\end{equation}

In this model, \( \mathbf{y} \) represents the observed values, \( \kappa(X, X) \) is the covariance matrix from the kernel function applied to all training input pairs, \( \epsilon^2 \) denotes noise variance, and \( I \) is the identity matrix. \( \kappa(X, x^*) \) and \( \kappa(x^*, X) \) are the covariances between training inputs and the test input.

Predictions for a new test point \( x^* \) are derived by conditioning this joint Gaussian distribution on the observed data.

\begin{figure}[!htb]
\centering
\includegraphics[width=0.5\textwidth]{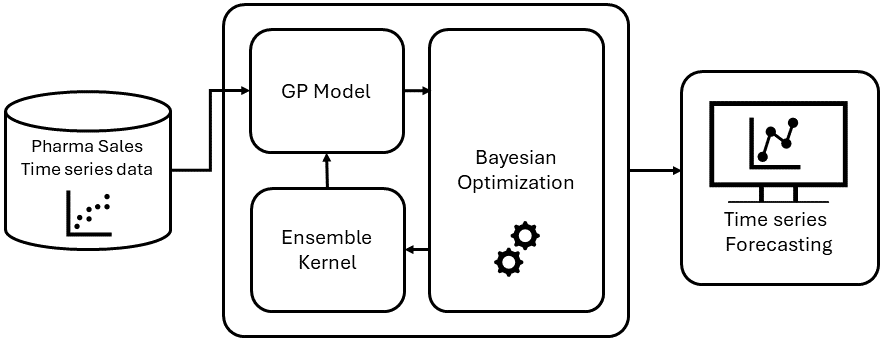}
\caption{Time series forcastion with an ensemble kernel Gaussian Process Regression approach}
\label{fig_n}
\end{figure}

\subsection{Insight into Gaussian Process Kernels}

The effectiveness of a Gaussian Process (GP) in modeling depends significantly on its kernel, also known as the covariance function. These kernels define crucial characteristics like the function's smoothness and variability. This section delves into three prevalent kernels: the Exponential Squared (ES) kernel, the Matérn kernel, and the Rational Quadratic (RQ) kernel, adapting to the revised notation.

\subsubsection{Exponential Squared (ES) Kernel}
Previously referred to as the SE kernel, the ES kernel is formulated as:
\begin{equation}
\kappa_{ES}(x, x') = \alpha^2 \exp\left(-\frac{\|x - x'\|^2}{2\lambda^2}\right)
\end{equation}
where \( \alpha^2 \) represents the variance and \( \lambda \) is the length-scale. Its infinite differentiability renders it exceptionally smooth and is thus favored for its simplicity and elegant properties.

\subsubsection{Revised Matérn Kernel}
The Matérn kernel, under the new notation, is expressed as:
\begin{equation}
\begin{split}
\kappa_{\text{Matérn}}(x, x') = \alpha^2 \frac{2^{1-\nu}}{\Gamma(\nu)}\left(\frac{\sqrt{2\nu}\|x - x'\|}{\lambda}\right)^\nu \\
\times K_\nu\left(\frac{\sqrt{2\nu}\|x - x'\|}{\lambda}\right)
\end{split}
\end{equation}

In this formulation, \( \alpha^2 \) is the variance, \( \lambda \) is the length-scale, and \( \nu \) is the smoothness parameter. The Matérn kernel provides a flexible smoothness control, making it a generalization of the ES kernel.

\subsubsection{Rational Quadratic (RQ) Kernel}
The RQ kernel, under the updated notation, is written as:
\begin{equation}
\kappa_{RQ}(x, x') = \alpha^2 \left(1 + \frac{\|x - x'\|^2}{2\beta \lambda^2}\right)^{-\beta}
\end{equation}
Here, \( \alpha^2 \) is the variance, \( \lambda \) the length-scale, and \( \beta \) a parameter influencing the mixture of scales. The RQ kernel is valuable for modeling functions with heterogeneous levels of smoothness.

Each kernel encodes distinct assumptions about the function and influences the learning and generalization capacity of the GP. Their selection hinges on the data's nature and the desired characteristics of the function to be modeled.

\subsection{Ensemble Kernel Method in Gaussian Processes}

The Ensemble Kernel Method in Gaussian Processes (GPs) combines multiple kernels to create a more robust and flexible model. This approach allows for capturing a broader range of features in the data. The ensemble kernel, \( \kappa_{ensemble}(x, x') \), is formulated by combining different kernels, each with their unique characteristics. A common ensemble approach is the linear combination of kernels:

\begin{equation}
\kappa_{ensemble}(x, x') = \sum_{i=1}^{N} w_i \kappa_i(x, x')
\end{equation}

where \( N \) is the number of kernels in the ensemble, \( w_i \) are the weights assigned to each kernel \( \kappa_i(x, x') \), and each \( \kappa_i \) is a distinct kernel like the Exponential Squared (ES), Matérn, or Rational Quadratic (RQ) kernel. The weights \( w_i \) are often learned from data, allowing the model to adaptively emphasize different characteristics captured by each kernel.

To optimize the ensemble kernel, parameters of individual kernels and the ensemble weights or powers are typically tuned using Bayesian Optimization with Gaussian Process priors. This method involves constructing a probabilistic model of the objective function and using it to select the most promising parameters to evaluate in the real world, balancing the trade-off between exploration and exploitation.

The Ensemble Kernel Method enhances the adaptability and expressiveness of GPs, making it suitable for sophisticated modeling tasks, especially in scenarios where data exhibits a mix of different behaviors or patterns.

\subsection{Bayesian Optimization with Gaussian Process Priors}

Bayesian Optimization (BO) is a strategy for the global optimization of black-box functions that are expensive to evaluate, particularly suited for hyperparameter tuning in machine learning. The method employs a Gaussian Process (GP) as a probabilistic model to estimate the unknown objective function.

\subsubsection{Gaussian Process as a Surrogate Model}
A GP acts as a surrogate model in BO. It is defined by Eq \ref{eq:gp_model}

\subsubsection{Bayesian Update and Acquisition Function}
The GP is updated as new evaluations are observed, forming a posterior distribution. The next sampling point is chosen using an acquisition function \( a(x) \), which balances exploration and exploitation. A common acquisition function is the Expected Improvement:
\begin{equation}
EI(x) = \mathbb{E}\left[\max(f(x) - f(x^+), 0)\right]
\end{equation}
where \( f(x^+) \) is the current best observation. The optimization of the ensemble kernel using the BO Algorithm is detailed in Algorithm 1.

\subsubsection{Advantages and Applications}
This optimization approach is highly effective in scenarios with expensive evaluations, scarce data, or complex high-dimensional spaces. Its applications range from tuning machine learning models to optimizing processes in material science and pharmaceuticals.

Implementing BO with GP priors requires careful selection of the kernel and acquisition function, ensuring effective exploration of the parameter space and convergence to optimal solutions.

\begin{algorithm}
\caption{Optimization of Kernel Weight in Gaussian Process}
\begin{algorithmic}[1]
\Procedure{OptimizeKernelWeight}{$Data, Model, Range, Iterations$}
    \State Initialize Bayesian Optimization: $BayesOptGP$
    \State $best~\omega \gets$ null
    \State $maxScore \gets -\infty$
    \For{$i \gets 1$ to $Iterations$}
        \State $\omega \gets$ AcquireThreshold($BayesOptGP$, $Range$)
        \State $score \gets$ EvaluateModel($Data$, $Model$, $\omega$)
        \If{$score > maxScore$}
            \State $maxScore \gets score$
            \State $best~\omega \gets \omega$
        \EndIf
        \State UpdateBayesOptGP($BayesOptGP$, $\omega$, $score$)
    \EndFor
    \State \textbf{return} $best~\omega$
\EndProcedure
\end{algorithmic}
\end{algorithm}

Fig \ref{fig_n} illustrates the entire process of time series forecasting. Initially, data is fed into the Gaussian Process (GP) model, followed by the application of the ensemble kernel to this model. The weights of the kernels are then optimized using Bayesian optimization, aiming to minimize the error and effectively determine the optimal time series.

\subsection{Evaluation Metrics}
To ensure accurate and reliable predictions from our Gaussian Process models, we utilize three fundamental metrics: Mean Squared Error (MSE), Mean Absolute Error (MAE), and the \(R^2\) score. The MSE, represented by \(\text{MSE} = \frac{1}{n}\sum_{i=1}^{n}(y_i - \hat{y}_i)^2\), quantifies the average squared difference between actual and estimated values, with a lower value indicating a better fit. The MAE, given by \(\text{MAE} = \frac{1}{n}\sum_{i=1}^{n}|y_i - \hat{y}_i|\), computes the average absolute difference and is less influenced by outliers. Lastly, the \(R^2\) score or coefficient of determination, expressed as \(R^2 = 1 - \frac{\text{SS}_{\text{res}}}{\text{SS}_{\text{tot}}}\), measures the variance proportion that's predictable, with a score closer to 1 denoting an excellent fit. Collectively, these metrics offer insights into the model's performance, guiding potential refinements and ensuring prediction dependability.

\section{Simulation \& Results}

\subsection{Categorization and Time Series Analysis of Pharmaceutical Sales Data} 

\begin{figure}[!htb]
\centering
\includegraphics[width=0.5\textwidth]{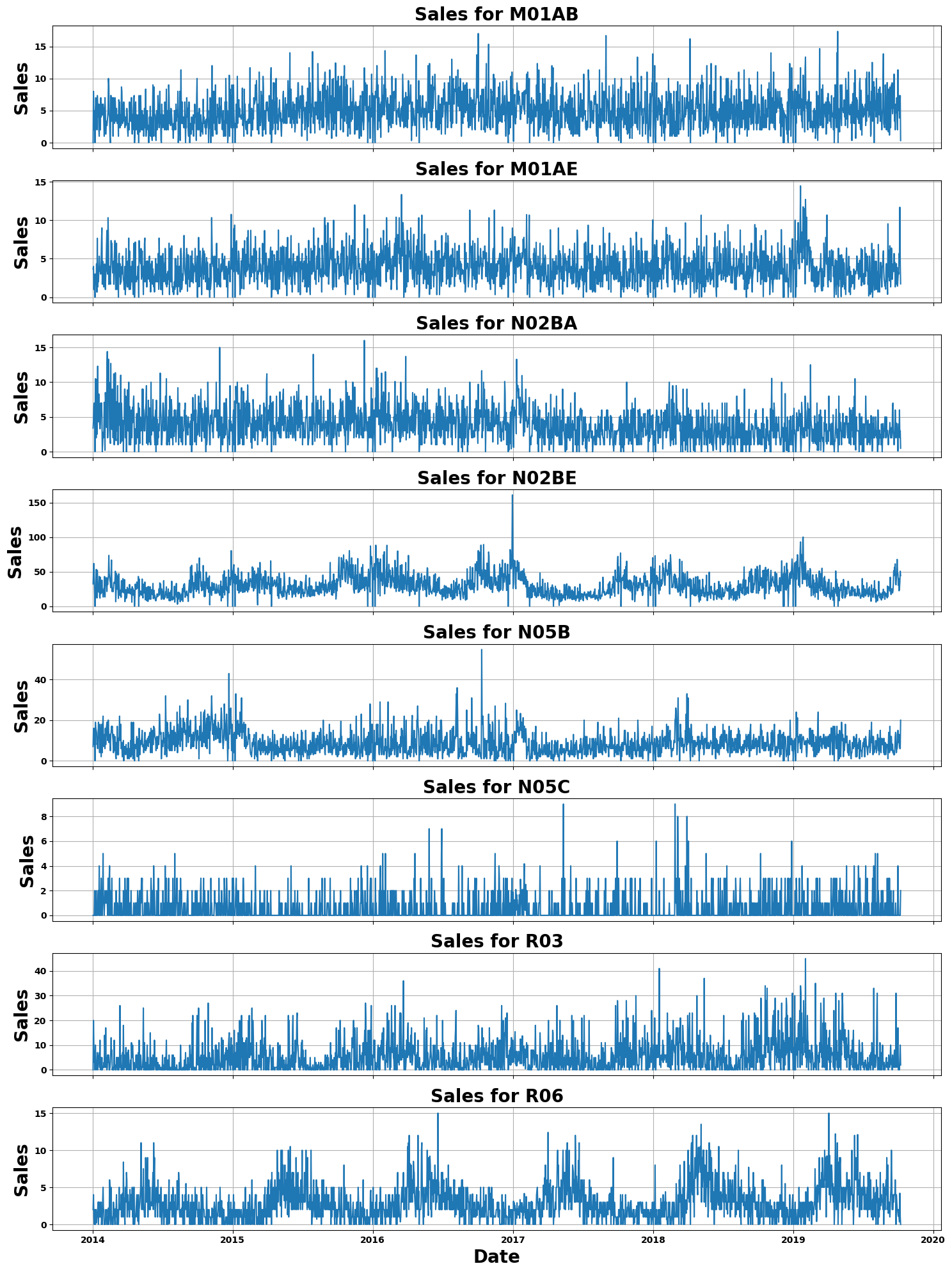}
\caption{Time Series Analysis of Pharmaceutical Sales by ATC Categories}
\label{fig_b}
\end{figure}

The initial dataset\cite{17}, comprising 600,000 transactional records from 2014 to 2019, includes details like the sale date, time, pharmaceutical brand, and quantity. Based on pharmacists' insights, the study pivoted from individual drugs to drug categories for better analysis and forecasting. This led to the classification of 57 drugs into 8 ATC categories: M01AB (non-steroid anti-inflammatory and antirheumatic products, Acetic acid derivatives), M01AE (non-steroid anti-inflammatory and antirheumatic products, Propionic acid derivatives), N02BA (other analgesics and antipyretics, Salicylic acid and derivatives), N02BE/B (other analgesics and antipyretics, Pyrazolones and Anilides), N05B (psycholeptic drugs, Anxiolytic drugs), N05C (psycholeptic drugs, Hypnotics and sedatives), R03 (drugs for obstructive airway diseases), and R06 (Antihistamines for systemic use). Time series analysis of these categories is illustrated in \ref{fig_b}.
In our study, we focused on analyzing the model based on GPR with various kernels, selecting the M01AB category (non-steroid anti-inflammatory and antirheumatic products, Acetic acid derivatives) for a detailed study. To ensure a comprehensive analysis, we employed a high-volume sampling method from this category. This approach involved randomly selecting a substantial number of samples from the dataset, ensuring a representative subset that captures the diverse trends and patterns within the M01AB category. By using high-volume sampling, we aimed to provide a robust and reliable analysis, minimizing the potential for sampling bias and ensuring that our findings are reflective of the broader trends in this pharmaceutical category.

\begin{figure}[!htb]
\centering
\includegraphics[width=0.5\textwidth]{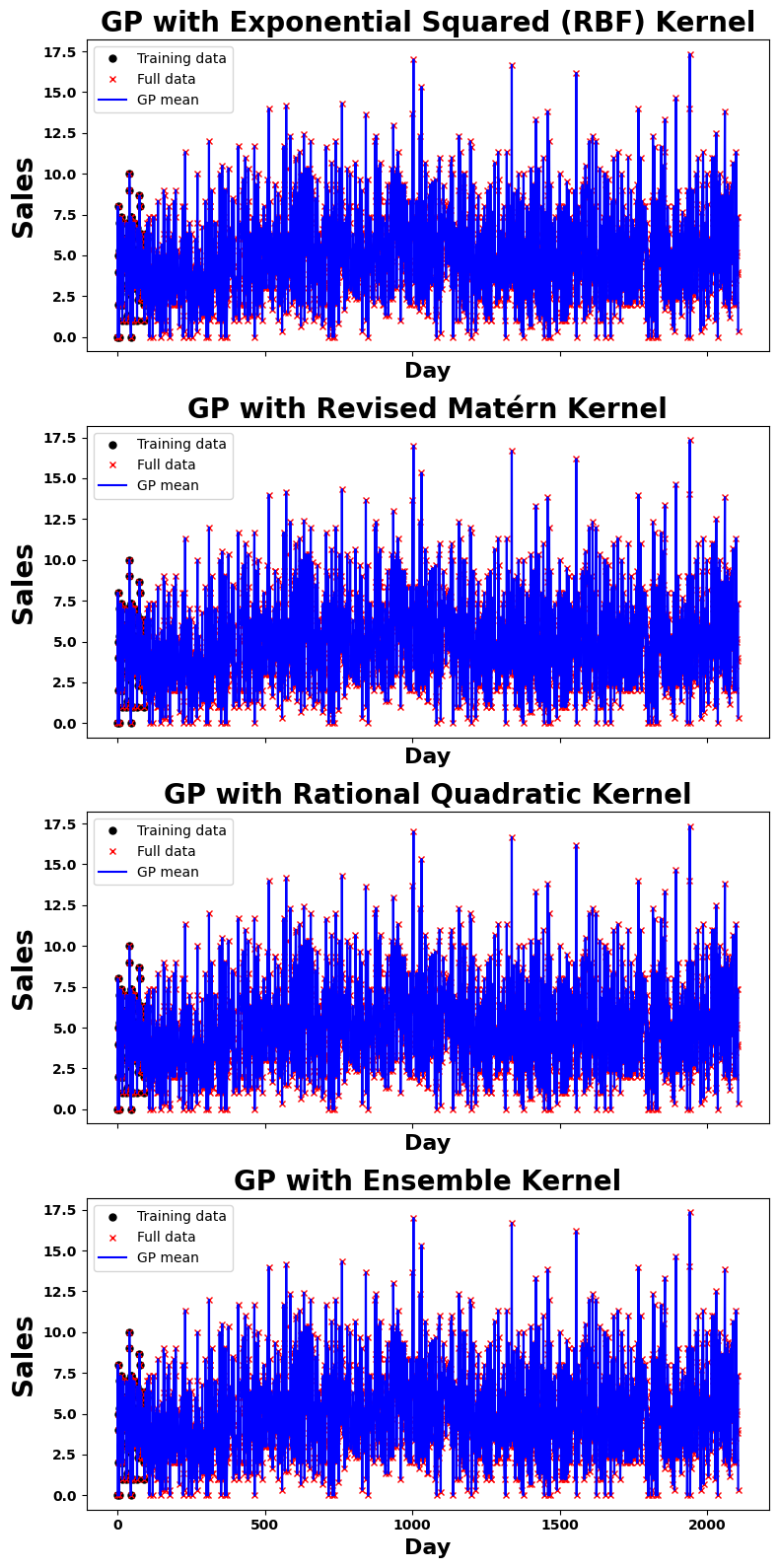}
\caption{Gaussian Process Regression Analysis Using RBF, Revised Matern, Rational Quadratic Kernels, and Their Ensemble}
\label{fig_c}
\end{figure}

In Fig \ref{fig_c}, we showcase the application of GPR with four distinct kernels—Exponential Squared, Revised Matern, Rational Quadratic, and an ensemble of these kernels—on samples from the M01AB category. Bayesian optimization was employed to determine the optimal weights for combining these kernels. The choice of these kernels was strategic: the RBF kernel is known for its smoothness and flexibility in modeling data; the Revised Matern kernel offers control over the smoothness of the function, making it versatile; the Rational Quadratic kernel can model varied levels of smoothness within the data. The ensemble approach combines these strengths, aiming to capture a comprehensive range of patterns and trends in the data, thus enhancing the model’s predictive power and robustness.

In the Bayesian optimization of our ensemble kernel, the assigned weights($\omega$) were 0.66 for the RBF kernel, 0.21 for the Revised Matern, and 0.13 for the Rational Quadratic kernel. The predominance of the Exponential Squared kernel underscores its effectiveness in modeling the core patterns in the data. The inclusion of the Revised Matern and Rational Quadratic kernels, albeit with lower weights, is crucial for their ability to handle data with varying smoothness and scale. Importantly, this ensemble approach significantly contributes to capturing and quantifying the uncertainty inherent in pharmaceutical sales data, enhancing the model's overall predictive accuracy and robustness in facing diverse data scenarios.

\begin{figure}[!htb]
\centering
\includegraphics[width=0.5\textwidth]{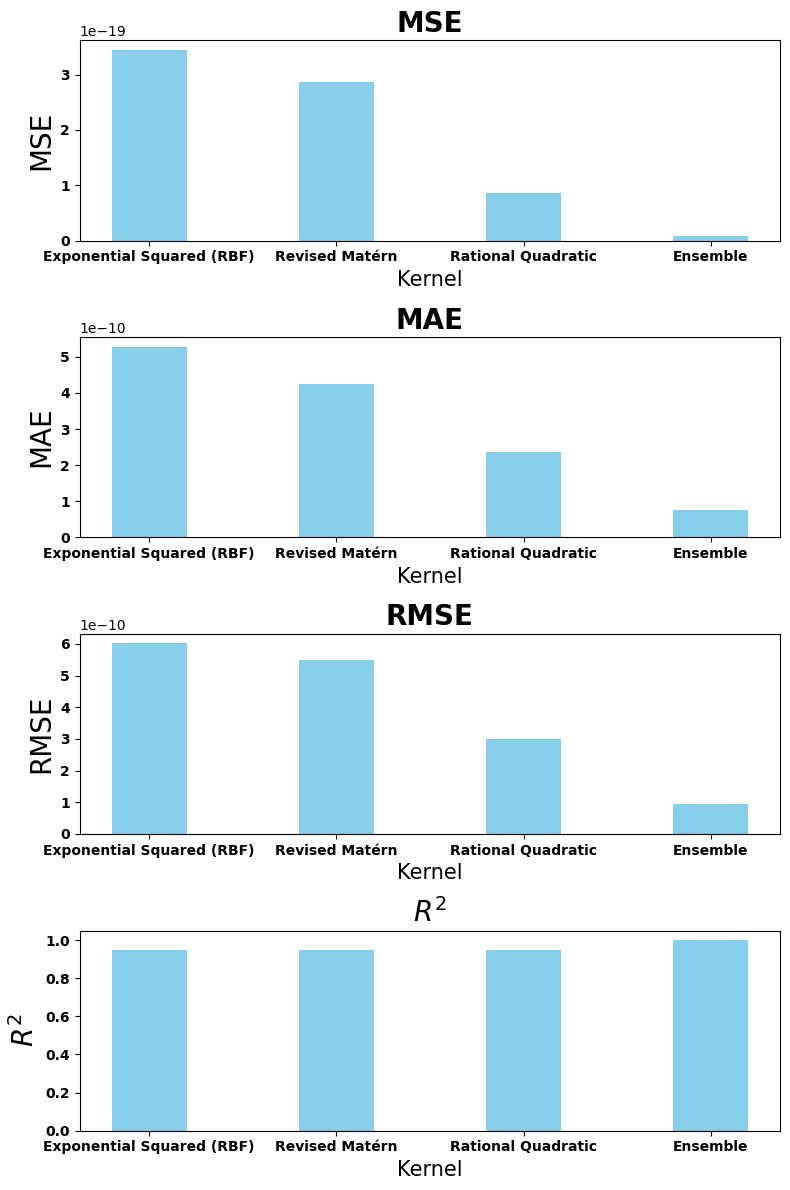}
\caption{Comparative Analysis of GPR Kernels Using MSE, MAE, RMSE, and R\textsuperscript{2} Metrics}
\label{fig_d}
\end{figure}

In Fig \ref{fig_d}, we analyze the performance of various kernels in Gaussian Process Regression (GPR) using metrics such as MSE, MAE, RMSE, and R-squared (R²). The Ensemble kernel exhibits exceptional performance, achieving an MSE of $9.09 \times 10^{-21}$, MAE of $7.48 \times 10^{-11}$, RMSE of $9.53 \times 10^{-11}$, and a perfect R² score of 1.0. In comparison, the Exponential Squared kernel records an MSE of $3.45 \times 10^{-19}$, MAE of $5.29 \times 10^{-10}$, and RMSE of $6.02 \times 10^{-10}$, with an R² of 0.95. Similar trends are observed with the Revised Matérn and Rational Quadratic kernels, where the Revised Matérn shows an MSE of $2.87 \times 10^{-19}$, MAE of $4.26 \times 10^{-10}$, RMSE of $5.49 \times 10^{-10}$, and the Rational Quadratic has an MSE of $8.59 \times 10^{-20}$, MAE of $2.35 \times 10^{-10}$, RMSE of $3.00 \times 10^{-10}$, both with R² scores of 0.95. These results clearly demonstrate the superior accuracy of the Ensemble kernel in our GPR model.

\section{conclusion}
This study demonstrates the potential of ensemble kernels in Gaussian Process Regression for enhancing predictive analytics in pharmaceutical sales. The integration of Exponential Squared, Revised Matérn, and Rational Quadratic kernels, optimized through Bayesian techniques, led to a model with remarkable accuracy, as evidenced by its performance metrics. Our approach not only addresses the complexities inherent in pharmaceutical sales data but also sets a precedent for future research in advanced time series analysis using ensemble kernels in GPR. This methodology could be pivotal in refining predictive models in various data-intensive fields.

\ifCLASSOPTIONcaptionsoff
  \newpage
\fi

% \end{IEEEbiography}


%\newpage


\end{document}